\documentclass[conference]{IEEEtran}
\IEEEoverridecommandlockouts

\usepackage{cite}
\usepackage{amsmath,amssymb}
\usepackage{graphicx}
\usepackage{array}
\usepackage{multirow}
\usepackage{tabularx}
\usepackage[caption=false,font=footnotesize]{subfig}
\usepackage[ruled,vlined,linesnumbered]{algorithm2e}
\usepackage{url}
\usepackage{balance}
\usepackage{placeins}
\usepackage{flushend}
\usepackage{microtype}
\usepackage{float}
\usepackage{xcolor}

\SetAlgoNlRelativeSize{-1}
\SetAlFnt{\footnotesize}
\SetAlCapFnt{\footnotesize}
\SetAlCapNameFnt{\footnotesize}
\SetKwInOut{Input}{Input}
\SetKwInOut{Output}{Output}

\title{Encoder-Sharing Hierarchical Federated Multi-Task Learning for VANETs}

\author{\IEEEauthorblockN{M. Saeid HaghighiFard and Sinem Coleri}
\IEEEauthorblockA{Department of Electrical and Electronics Engineering, Ko\c{c} University, Istanbul, T\"urkiye\\
Email: \{mhaghighifard21, scoleri\}@ku.edu.tr}
\thanks{This work was supported by the Scientific and Technological Research Council of Turkey (T\"UB\.ITAK) under Grant 119C058 and Ford Otosan.}}

\begin{document}
\maketitle

\begin{abstract}
Most federated learning frameworks for vehicular ad hoc networks assume that all vehicles collaboratively train a single model for a common task. This assumption limits their applicability to practical vehicular environments, where vehicles may perform heterogeneous but related perception tasks with different output spaces. This paper proposes encoder-sharing hierarchical multi-task federated learning (EN-HMTFL), which integrates cluster-based hierarchical federated learning with a globally shared encoder and vehicle-local decoders. EN-HMTFL enables vehicles performing different tasks to collaboratively learn a transferable feature representation while preserving their task-specific models locally. Only the encoder is exchanged and aggregated through the hierarchy, whereas raw data and local decoder parameters remain at the vehicles. The proposed framework is evaluated on the MNIST and GTSRB datasets in different vehicular scenarios. Across the evaluated scenarios, EN-HMTFL improves accuracy by up to 24.0\% relative to the compared representation-sharing benchmark. In scenarios where EN-HMTFL converges earlier, the reduction reaches up to 69 communication rounds (28.8\%). 
\end{abstract}


\section{Introduction}
\IEEEPARstart{V}{ehicular} Ad hoc Networks (VANETs) increasingly employ machine learning (ML) for safety- and mobility-critical services such as traffic-sign recognition, object detection, trajectory prediction, traffic-flow estimation, and driving-scene understanding. These services rely on the large volume of observations generated by onboard cameras, radar, LiDAR, and other sensors~\cite{zhang2024fl_its,wang2023flcav,chellapandi2024cav_survey}. A conventional centralized learning architecture requires vehicles to upload their locally generated data to a remote processor. Such an architecture becomes difficult to scale as the number of vehicles and the volume of sensory data increase; it also consumes radio resources and exposes raw observations outside the vehicle~\cite{lim2020fl_survey}. Federated learning (FL) addresses these limitations by allowing vehicles to train locally and exchange model parameters instead of raw data~\cite{mcmahan2017fedavg,elbir2022fl_vehicular}. However, most FL formulations assume that every participating vehicle optimizes the same model for one common learning objective.

Multi-task federated learning (MTFL) replaces the single-task assumption by coordinating multiple related learning objectives. The original federated multi-task formulation treats clients as related tasks and jointly optimizes their models while accounting for communication and systems constraints~\cite{smith2017fmtl}. FedEM learns a mixture of shared models and adapts the mixture coefficients to each client distribution~\cite{marfoq2021fedem}. Ditto jointly learns a global reference model and a personalized model for every client~\cite{li2021ditto}. When Ditto is executed independently for each task, it provides within-task personalization but does not create a common representation that enables heterogeneous tasks to collaborate. These methods, therefore, do not directly resolve the incompatibility among task-specific output components.

A natural solution is to divide each model into a representation component and a task-dependent component. FedRep exchanges a common representation while retaining personalized prediction layers~\cite{collins2021fedrep}. M-Fed similarly adopts an encoder-decoder architecture to support collaboration across heterogeneous tasks while preserving task-dependent components locally~\cite{cao2025mfed}. In this design, the encoder learns transferable features from vehicles performing different tasks, whereas the decoder maps these features to the output space required by each vehicle and task. However, existing representation-sharing methods generally rely on a flat client-server architecture, in which every participating vehicle sends its model update directly to a central server and receives the updated global representation directly from that server in each communication round. In VANETs, this direct exchange can overload the infrastructure link, scale poorly with the number of vehicles, and fail to exploit the temporary local connectivity among nearby vehicles.

This paper proposes an encoder-sharing hierarchical multi-task federated learning framework for cluster-based vehicular networks. Within the proposed framework, vehicles performing heterogeneous but related tasks jointly learn a shared encoder through the hierarchy while retaining their task-specific decoders locally. We conduct extensive simulations under different vehicular network scenarios. Comparisons with two related benchmarks, Ditto and M-Fed, demonstrate that the proposed algorithm achieves the highest accuracy across the evaluated scenarios and can reduce the number of communication rounds required for convergence, depending on the scenario.

The remainder of this paper is organized as follows. Section~II presents the system model, Section~III describes the proposed encoder-sharing hierarchical MTFL algorithm, Section~IV provides the experimental evaluation, and Section~V concludes the paper.

\section{System Model}
We consider a dynamic vehicular ad hoc network (VANET) in which vehicles communicate through vehicle-to-vehicle (V2V) links and access the Evolved Packet Core (EPC) through vehicle-to-infrastructure (V2I) or vehicle-to-network (V2N) connectivity. V2V communication may be supported by IEEE 802.11p~\cite{5888501}, IEEE 802.11bd~\cite{9779322}, or LTE-based device-to-device communication~\cite{7497762}, whereas V2I/V2N connectivity is provided through 5G New Radio V2X~\cite{9392787}.

The cluster-based hierarchical federated learning (CbHFL) architecture introduced in~\cite{haghighifard2025hfl} serves as the underlying communication and coordination framework. In hierarchical federated learning (HFL), intermediate aggregators are placed between participating clients and the central server so that local model exchanges are first collected and aggregated at an intermediate level before being forwarded to the central entity. This hierarchy reduces the number of direct client-server transmissions and localizes frequent model exchanges.

In CbHFL, vehicles dynamically transition among four states: \emph{INITIAL} (IN), \emph{STATE ELECTION} (SE), \emph{CLUSTER HEAD} (CH), and \emph{CLUSTER MEMBER} (CM). Each vehicle maintains a Vehicle Information Base (VIB) containing its current state, mobility information, neighboring vehicles, candidate CHs, active task set, shared encoder parameters, and task-specific local decoder parameters. The VIB is continuously updated through state transitions and periodic ``HELLO\_PACKET'' exchanges, enabling distributed cluster formation and maintenance under dynamic network conditions.

After initialization, every vehicle enters the SE state and attempts to associate with a neighboring CH. When multiple candidate CHs are available, the vehicle associates with the CH that offers the most stable mobility relationship. If no suitable CH is available, a CH election is performed among neighboring vehicles in the SE state, selecting the vehicle with the lowest average relative speed to its neighbors as the CH; otherwise, the vehicle remains in the SE state until the topology changes. During network operation, cluster membership is continuously updated: CMs that lose connectivity re-enter the SE state, whereas neighboring CHs may merge to eliminate redundant clusters and reduce communication toward the EPC. 

The resulting hierarchy consists of three entities: CMs, CHs, and the EPC. During global round $r$, let $\mathcal{C}^{(r)}$ denote the set of active clusters and $\mathcal{S}_{c}^{(r)}$ denote the set of CMs associated with CH $c$ that successfully complete local training. Each CM trains its local multi-task model using private data and uploads only its shared encoder parameters. Each CH $c$ aggregates the encoder parameters received from $\mathcal{S}_{c}^{(r)}$ into a cluster-level encoder and forwards it to the EPC, which performs inter-cluster aggregation to construct the global shared encoder. The updated encoder is then disseminated through the reverse hierarchy. Since task-specific decoders remain local throughout the learning process, they are never exchanged, aggregated, or involved in cluster formation.

\section{Encoder-Sharing Hierarchical MTFL}
The proposed algorithm has three successive components. First, each CM jointly trains the shared encoder with the decoder for every locally supported task. Second, each CH averages only the encoders received from its CMs and forwards the cluster encoder to the EPC. Third, the EPC averages the cluster encoders, redistributes the resulting global encoder, and checks convergence using the EPC-level accuracy. The decoder parameters do not participate in either hierarchical average.

\FloatBarrier
\begin{algorithm}[ht]
\caption{Local Training at Cluster Member (CM) with Shared Encoder and Local Decoders}
\label{alg:cm}
\Input{Global encoder $\boldsymbol{\theta}^{\mathrm{enc},(r)}_{\mathrm{EPC}}$; task set $\mathcal{T}_{i}$; local datasets $\{\mathcal{D}_{i,t}\}_{t\in\mathcal{T}_{i}}$; local decoders $\{\boldsymbol{\theta}^{\mathrm{dec}}_{i,t}\}_{t\in\mathcal{T}_{i}}$; local epochs $E_i$}
\Output{Updated encoder $\boldsymbol{\theta}^{\mathrm{enc},(r)}_{i}$; updated local decoders retained at vehicle $i$}
$\boldsymbol{\theta}^{\mathrm{enc}}_{i}\leftarrow\boldsymbol{\theta}^{\mathrm{enc},(r)}_{\mathrm{EPC}}$\;
Retain $\{\boldsymbol{\theta}^{\mathrm{dec}}_{i,t}\}_{t\in\mathcal{T}_{i}}$ from the preceding round\;
\For{$e=1,\ldots,E_i$}{
$L_i\leftarrow0$\;
\ForEach{$t\in\mathcal{T}_{i}$}{
Sample mini-batch $\mathcal{B}_{i,t}\subset\mathcal{D}_{i,t}$\;
$\mathbf{z}_{i,t}\leftarrow\mathrm{Enc}(x_{i,t};\boldsymbol{\theta}^{\mathrm{enc}}_{i})$\;
$\widehat{y}_{i,t}\leftarrow\mathrm{Dec}_{i,t}(\mathbf{z}_{i,t};\boldsymbol{\theta}^{\mathrm{dec}}_{i,t})$\;
$L_{i,t}\leftarrow\ell_t(\widehat{y}_{i,t},y_{i,t})$\;
$L_i\leftarrow L_i+L_{i,t}$\;
}
$\boldsymbol{\theta}^{\mathrm{enc}}_{i}\leftarrow\boldsymbol{\theta}^{\mathrm{enc}}_{i}-\eta_{\mathrm{enc}}\nabla_{\boldsymbol{\theta}^{\mathrm{enc}}_{i}}L_i$\;
\ForEach{$t\in\mathcal{T}_{i}$}{
$\boldsymbol{\theta}^{\mathrm{dec}}_{i,t}\leftarrow\boldsymbol{\theta}^{\mathrm{dec}}_{i,t}-\eta_t\nabla_{\boldsymbol{\theta}^{\mathrm{dec}}_{i,t}}L_{i,t}$\;
}
}
$\boldsymbol{\theta}^{\mathrm{enc},(r)}_{i}\leftarrow\boldsymbol{\theta}^{\mathrm{enc}}_{i}$ and send it to the associated CH\;
Keep all $\{\boldsymbol{\theta}^{\mathrm{dec}}_{i,t}\}_{t\in\mathcal{T}_{i}}$ at vehicle $i$\;
\end{algorithm}

Algorithm~\ref{alg:cm} describes the local multi-task learning procedure executed at each CM. Let $\mathcal{T}$ denote the set of learning tasks considered by the system, and let $\mathcal{T}_{i}\subseteq\mathcal{T}$ denote the nonempty subset of tasks supported by vehicle $i$. Each task $t\in\mathcal{T}$ has its own input space $\mathcal{X}_{t}$, output space $\mathcal{Y}_{t}$, and loss function $\ell_t$. For every supported task $t\in\mathcal{T}_{i}$, vehicle $i$ stores a private non-IID dataset
\begin{equation}
\mathcal{D}_{i,t}=\{(x_{i,t}^{(n)},y_{i,t}^{(n)})\}_{n=1}^{N_{i,t}},
\label{eq:dataset}
\end{equation}
where $N_{i,t}$ denotes the number of local samples of task $t$ available at vehicle $i$. The dataset is never transmitted to another vehicle, a CH, or the EPC. Vehicle $i$ maintains a shared encoder $\mathrm{Enc}(\cdot;\boldsymbol{\theta}^{\mathrm{enc}}_{i})$ and a task-specific decoder $\mathrm{Dec}_{i,t}(\cdot;\boldsymbol{\theta}^{\mathrm{dec}}_{i,t})$ for every $t\in\mathcal{T}_{i}$. Here, $\boldsymbol{\theta}^{\mathrm{enc}}_{i}$ denotes the shared encoder parameters at vehicle $i$, and $\boldsymbol{\theta}^{\mathrm{dec}}_{i,t}$ denotes the local decoder parameters for task $t$ at vehicle $i$. For a task-specific input $x_{i,t}$, the encoder first produces the latent representation
\begin{align}
\mathbf{z}_{i,t}&=\mathrm{Enc}(x_{i,t};\boldsymbol{\theta}^{\mathrm{enc}}_{i}),
\label{eq:encode}\\
\widehat{y}_{i,t}&=\mathrm{Dec}_{i,t}(\mathbf{z}_{i,t};\boldsymbol{\theta}^{\mathrm{dec}}_{i,t}),
\label{eq:decode}
\end{align}
where $\mathbf{z}_{i,t}$ is the latent representation generated for task $t$ at vehicle $i$, and $\widehat{y}_{i,t}$ is the corresponding predicted output produced by the task-specific decoder, which maps the shared latent representation to the output space of task $t$. The shared encoder has the same architecture and parameter dimensions across vehicles, enabling its parameters to be aggregated hierarchically. The encoder parameters are the only trainable parameters exchanged through the hierarchical architecture, whereas decoder parameters are both vehicle- and task-specific and always remain local. The corresponding task loss and local multi-task objective are defined as
\begin{align}
L_{i,t}&=\ell_t(\widehat{y}_{i,t},y_{i,t}), \qquad t\in\mathcal{T}_{i},
\label{eq:taskloss}\\
L_i&=\sum_{t\in\mathcal{T}_{i}}L_{i,t}.
\label{eq:totalloss}
\end{align}

At the beginning of each communication round $r$, every participating CM replaces its local encoder $\boldsymbol{\theta}^{\mathrm{enc}}_{i}$ with the latest global encoder $\boldsymbol{\theta}^{\mathrm{enc},(r)}_{\mathrm{EPC}}$ received from the EPC while preserving all decoder parameters learned in previous rounds (Lines~1--2). For each local epoch, the CM initializes the total loss and iterates over every task in its assigned task set; consequently, a vehicle supporting multiple tasks jointly trains all corresponding decoders during the same communication round (Lines~3--5). For each task, the CM samples a mini-batch from its private dataset, computes the latent representation using the shared encoder, applies the corresponding decoder, evaluates the task loss, and accumulates it into the local multi-task objective (Lines~6--10). The shared encoder is then updated using the gradient of the complete multi-task objective, integrating knowledge learned from different tasks into a common representation, whereas each decoder is updated only with the loss of its corresponding task, preventing interference between task-specific output spaces (Lines~11--13). After local optimization, only the updated encoder parameters $\boldsymbol{\theta}^{\mathrm{enc},(r)}_{i}$ are transmitted to the associated CH, while all decoder parameters and private datasets remain local to the vehicle (Lines~14--15).

\FloatBarrier
\begin{algorithm}[ht]
\caption{Intra-Cluster Encoder Aggregation at Cluster Head (CH)}
\label{alg:ch}
\Input{Encoder updates $\{\boldsymbol{\theta}^{\mathrm{enc},(r)}_{i}\}_{i\in\mathcal{S}_{c}^{(r)}}$ from the active CMs of cluster $c$}
\Output{Cluster encoder $\boldsymbol{\theta}^{\mathrm{enc},(r)}_{c}$; disseminated EPC encoder}
Collect $\boldsymbol{\theta}^{\mathrm{enc},(r)}_{i}$ from every $i\in\mathcal{S}_{c}^{(r)}$\;
$\boldsymbol{\theta}^{\mathrm{enc},(r)}_{c}\leftarrow\frac{1}{|\mathcal{S}_{c}^{(r)}|}\sum_{i\in\mathcal{S}_{c}^{(r)}}\boldsymbol{\theta}^{\mathrm{enc},(r)}_{i}$\;
Send $\boldsymbol{\theta}^{\mathrm{enc},(r)}_{c}$ to the EPC\;
Receive $\boldsymbol{\theta}^{\mathrm{enc},(r+1)}_{\mathrm{EPC}}$ from the EPC\;
Broadcast $\boldsymbol{\theta}^{\mathrm{enc},(r+1)}_{\mathrm{EPC}}$ to all associated CMs\;
\end{algorithm}
\FloatBarrier

Algorithm~\ref{alg:ch} first collects one encoder update from each active CM in the current cluster (Line~1). It then applies the average for cluster $c$. Decoder parameters and task identities are not inputs to this operation (Line~2). The CH sends the resulting cluster encoder to the EPC rather than forwarding every individual CM encoder over the infrastructure link (Line~3). After the EPC completes the next global update, the CH receives the new global encoder and broadcasts it to its CMs for the next round (Lines~4--5).

\FloatBarrier
\begin{algorithm}[ht]
\caption{Global Encoder Aggregation at EPC}
\label{alg:epc}
\Input{Initial global encoder $\boldsymbol{\theta}^{\mathrm{enc},(0)}_{\mathrm{EPC}}$}
\Output{Global encoder sequence}
\For{$r=0,\ldots,R_{\max}-1$}{
Collect $\{\boldsymbol{\theta}^{\mathrm{enc},(r)}_{c}\}_{c\in\mathcal{C}^{(r)}}$ from the active CHs\;
$\boldsymbol{\theta}^{\mathrm{enc},(r+1)}_{\mathrm{EPC}}\leftarrow\frac{1}{|\mathcal{C}^{(r)}|}\sum_{c\in\mathcal{C}^{(r)}}\boldsymbol{\theta}^{\mathrm{enc},(r)}_{c}$\;
Broadcast $\boldsymbol{\theta}^{\mathrm{enc},(r+1)}_{\mathrm{EPC}}$ to all active CHs\;}
\end{algorithm}
\FloatBarrier

For algorithm~\ref{alg:epc}, in every global round, the EPC collects one encoder from each active CH, computes the average, and sends the new global encoder back to the CHs.

\FloatBarrier
\section{Experimental Evaluation}

We compare the proposed encoder-sharing hierarchical multi-task federated learning algorithm, termed EN-HMTFL, with Ditto and M-Fed, as these benchmarks capture the two main learning paradigms most relevant to our design. Ditto is adopted as a personalized federated learning benchmark and adapted to the underlying vehicular setting by executing independent Ditto processes in parallel for the tasks under consideration~\cite {li2021ditto}. Each process learns a task-specific global reference model together with personalized local models for the vehicles assigned to that task. Hence, Ditto provides personalization within each task, but it does not enable knowledge transfer or representation sharing across heterogeneous tasks. M-Fed is selected as the closest representation-sharing benchmark because it adopts an encoder-decoder architecture and enables cross-task knowledge transfer through a global encoder~\cite{cao2025mfed}. However, M-Fed does not exchange only encoder parameters. Clients upload both their encoders and task-specific decoders, and the server first aggregates complete models belonging to the same task to construct task-global models. It then extracts and aggregates the encoder components of these task-global models to obtain a cross-task global encoder. In contrast, EN-HMTFL communicates and aggregates only encoder parameters through the hierarchy, while all decoders remain strictly local and are never transmitted or aggregated.

\subsection{Simulation Setup}

The mobility and communication environment is generated using SUMO, while Kafka supports communication events and model updates via streaming~\cite{SUMO,KAFKA}. Learning is implemented in PyTorch and scikit-learn. IEEE~802.11p with the Winner+ B1 channel model is used for V2V communication, and the infrastructure link follows the Friis-based 5G NR model~\cite{sepulcre2022v2v,3gpptr38901}. The evaluated scenarios include a 50-vehicle setting with a 100~m transmission range and a 20-vehicle setting with transmission ranges of 100 and 500~m. Vehicles are randomly assigned to perform MNIST, GTSRB, or both tasks, and each vehicle trains using its own private non-IID dataset.

For GTSRB, a lightweight CNN extracts a 32-dimensional feature vector from each image, while MNIST samples are flattened into 64-dimensional vectors. These features are mapped by a three-layer shared encoder to a 128-dimensional latent representation. Each vehicle maintains a local reconstruction decoder and a task-specific classification head; only the encoder parameters are exchanged and aggregated. The transmitted encoder accounts for approximately 46.9\% of the MNIST model parameters and 45.7\% of the GTSRB model parameters.

Training uses SGD with momentum $0.9$, an initial learning rate of $0.01$, and a decay factor of $0.95$ every 10 communication rounds. Each CM performs 10 local epochs per round. Mean-squared error is used for reconstruction, and cross-entropy for classification. EPC accuracy is evaluated over 250 rounds, and convergence is declared when the change in accuracy remains below $0.005$ for three consecutive rounds. The reported convergence round is the first to complete this three-round streak.

\begin{figure*}[t]
\centering
\subfloat[MNIST]{\includegraphics[width=0.485\textwidth]{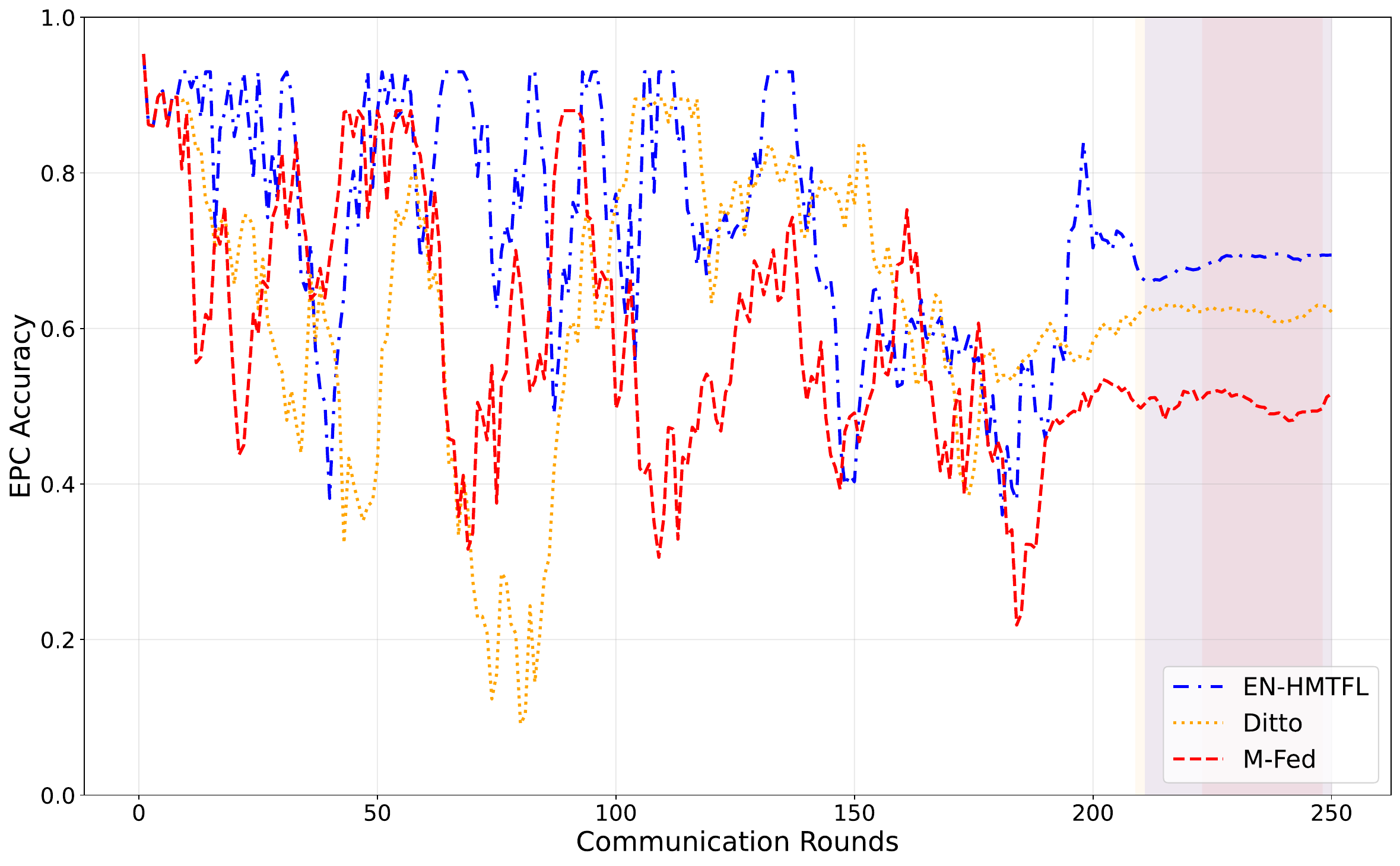}\label{fig:50_mnist}}
\hfill
\subfloat[GTSRB]{\includegraphics[width=0.485\textwidth]{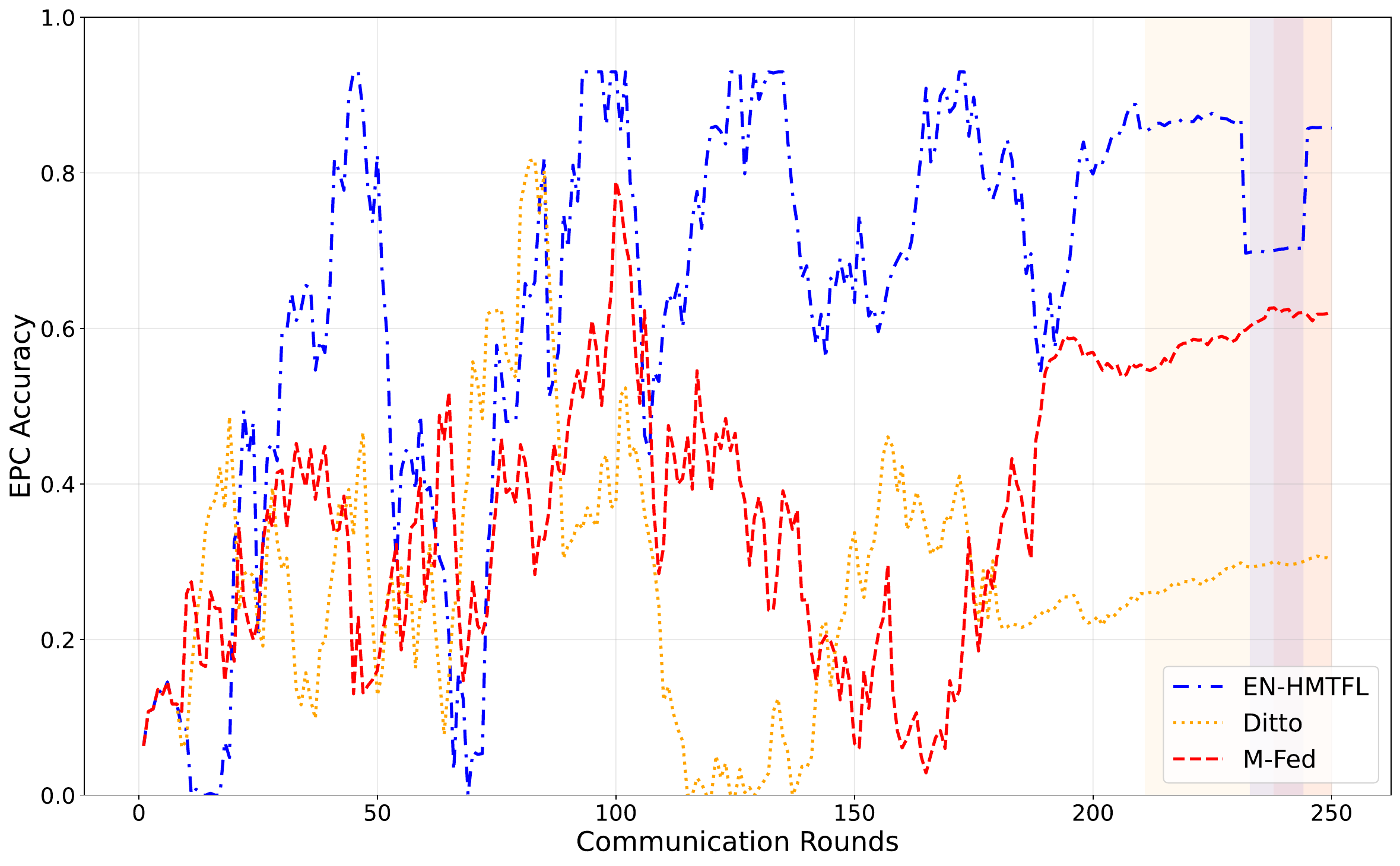}\label{fig:50_gtsrb}}
\caption{EPC accuracy versus communication round for 50 vehicles, a 100~m transmission range: (a) MNIST and (b) GTSRB.}
\label{fig:50veh}
\end{figure*}

\subsection{Performance evaluation}
Fig.~\ref{fig:50veh} compares EN-HMTFL with Ditto and M-Fed for 50 vehicles and a transmission range of 100~m. On MNIST, EN-HMTFL achieves a late-round mean EPC accuracy of $69.29\pm0.22\%$, outperforming Ditto and M-Fed, which attain $61.89\pm0.67\%$ and $49.81\pm1.06\%$, respectively. This corresponds to gains of 7.41 percentage points (12.0\% relative) over Ditto and 19.49 percentage points (39.1\% relative) over M-Fed. EN-HMTFL satisfies the convergence criterion at round 211, only two rounds after Ditto and 12 rounds before M-Fed. More importantly, its late-round standard deviation is substantially smaller than those of both benchmarks, indicating that the proposed method not only achieves higher accuracy but also maintains a more stable operating regime. Therefore, the slight two-round delay relative to Ditto is negligible compared with the considerable improvement in sustained accuracy and stability.

The performance gap becomes more pronounced on GTSRB. EN-HMTFL achieves a late-round mean EPC accuracy of $76.39\pm7.79\%$, compared with $29.92\pm0.41\%$ for Ditto and $61.58\pm0.86\%$ for M-Fed. Accordingly, EN-HMTFL improves the accuracy by 46.47 percentage points (155.3\% relative) over Ditto and by 14.81 percentage points (24.0\% relative) over M-Fed. Although Ditto meets the stopping criterion earlier, at round 211, this earlier stabilization occurs at a markedly inferior accuracy level and therefore does not indicate a better learned solution. EN-HMTFL converges at round 233, five rounds earlier than M-Fed, and reaches a final plotted accuracy of 85.74\%, whereas Ditto and M-Fed reach only 30.32\% and 62.25\%, respectively. The larger temporal variation of EN-HMTFL on GTSRB is caused by a temporary late-round degradation; however, the method subsequently recovers and remains clearly superior in both its final and sustained accuracy.

Overall, EN-HMTFL achieves the highest late-round EPC accuracy in the 50-vehicle scenario, while its convergence round is comparable to the benchmarks and depends on the task. Unlike Ditto, it enables cross-task representation sharing, and unlike M-Fed, it exploits localized CbHFL aggregation. This combination improves knowledge transfer and yields higher EPC accuracy, particularly for the more heterogeneous GTSRB task.

\begin{figure*}[t]
\centering
\subfloat[100~m, MNIST]{\includegraphics[width=0.475\textwidth]{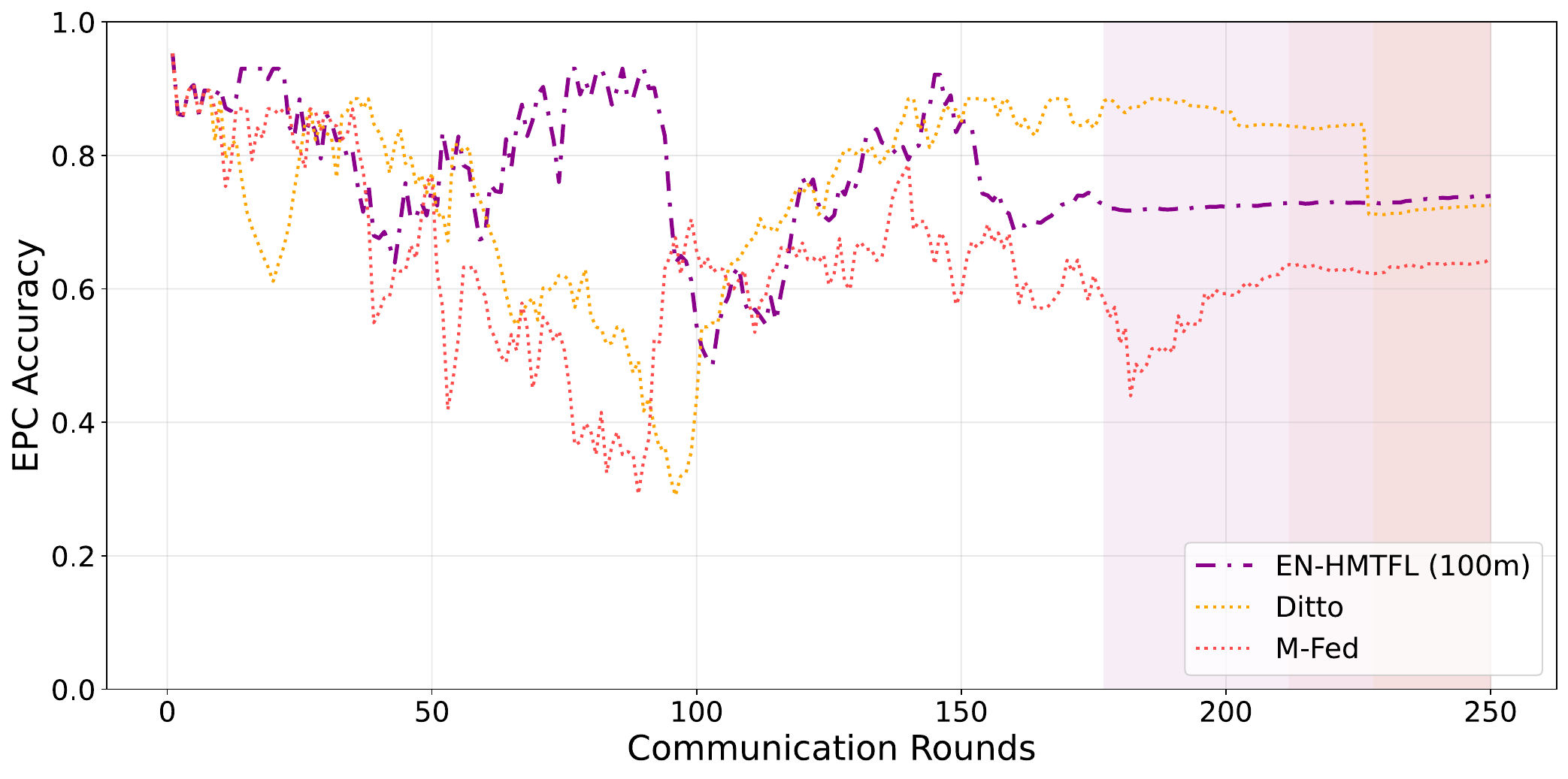}\label{fig:100_mnist}}
\hfill
\subfloat[100~m, GTSRB]{\includegraphics[width=0.475\textwidth]{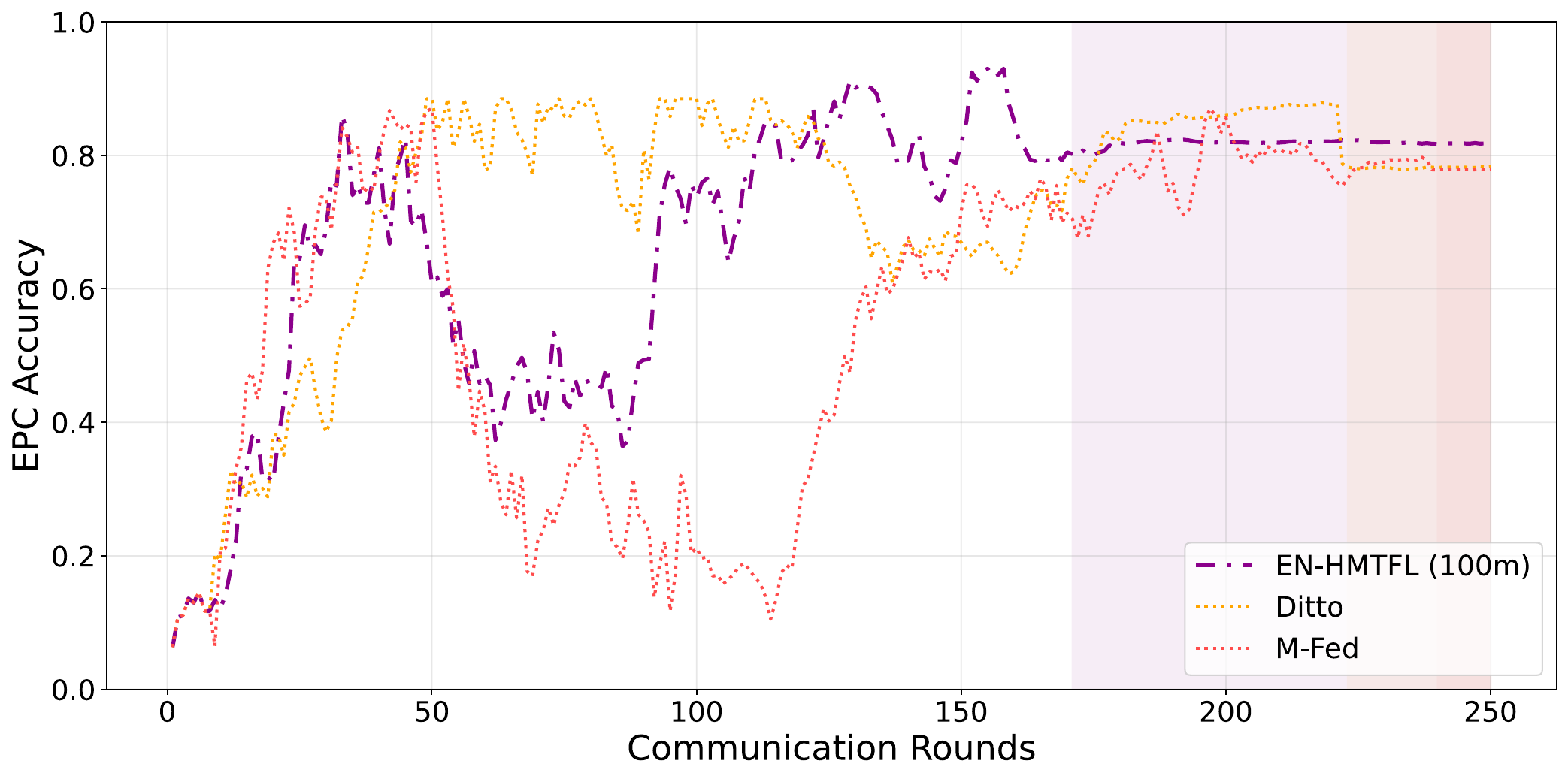}\label{fig:100_gtsrb}}\\[-0.5ex]
\subfloat[500~m, MNIST]{\includegraphics[width=0.475\textwidth]{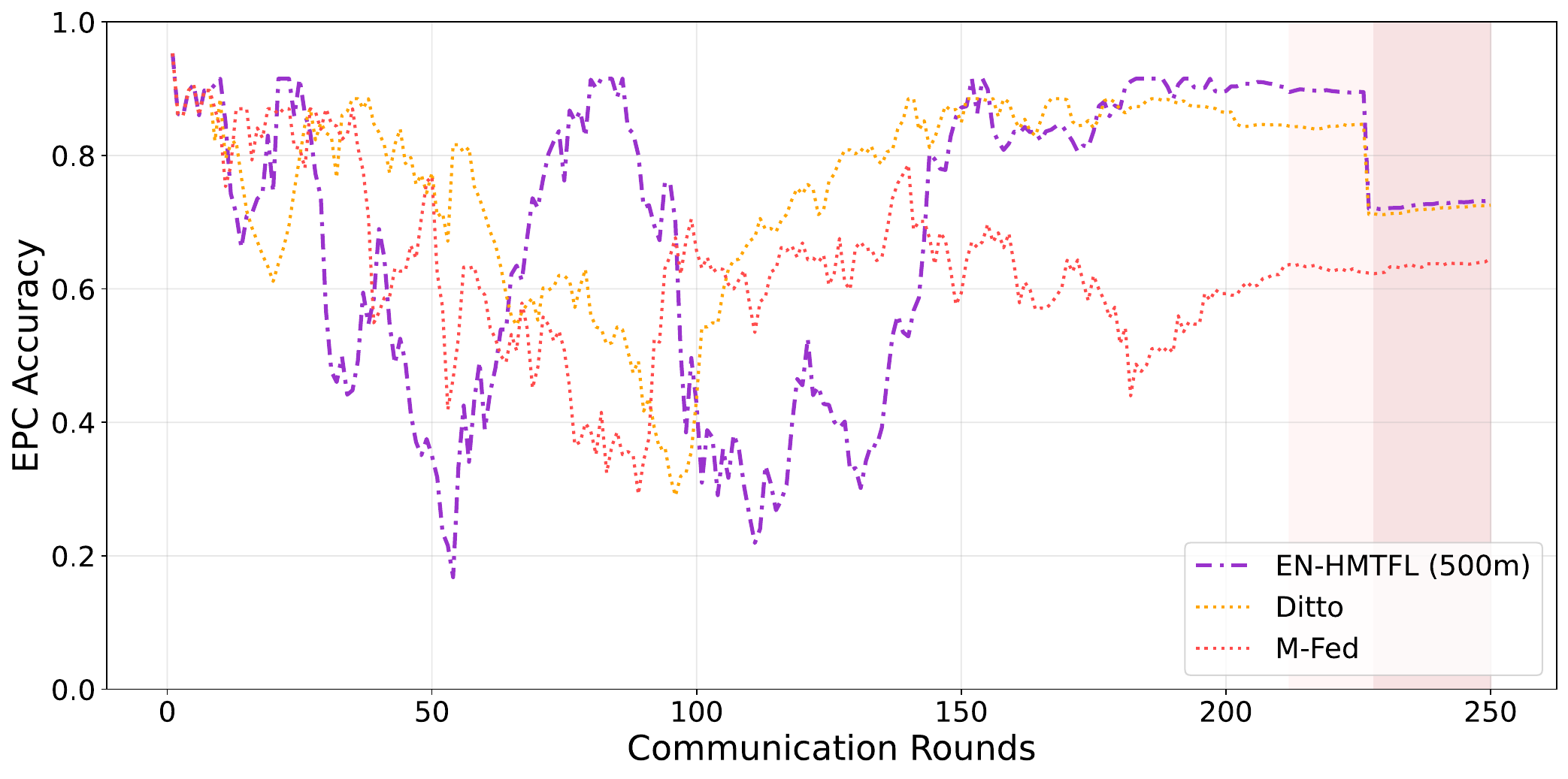}\label{fig:500_mnist}}
\hfill
\subfloat[500~m, GTSRB]{\includegraphics[width=0.475\textwidth]{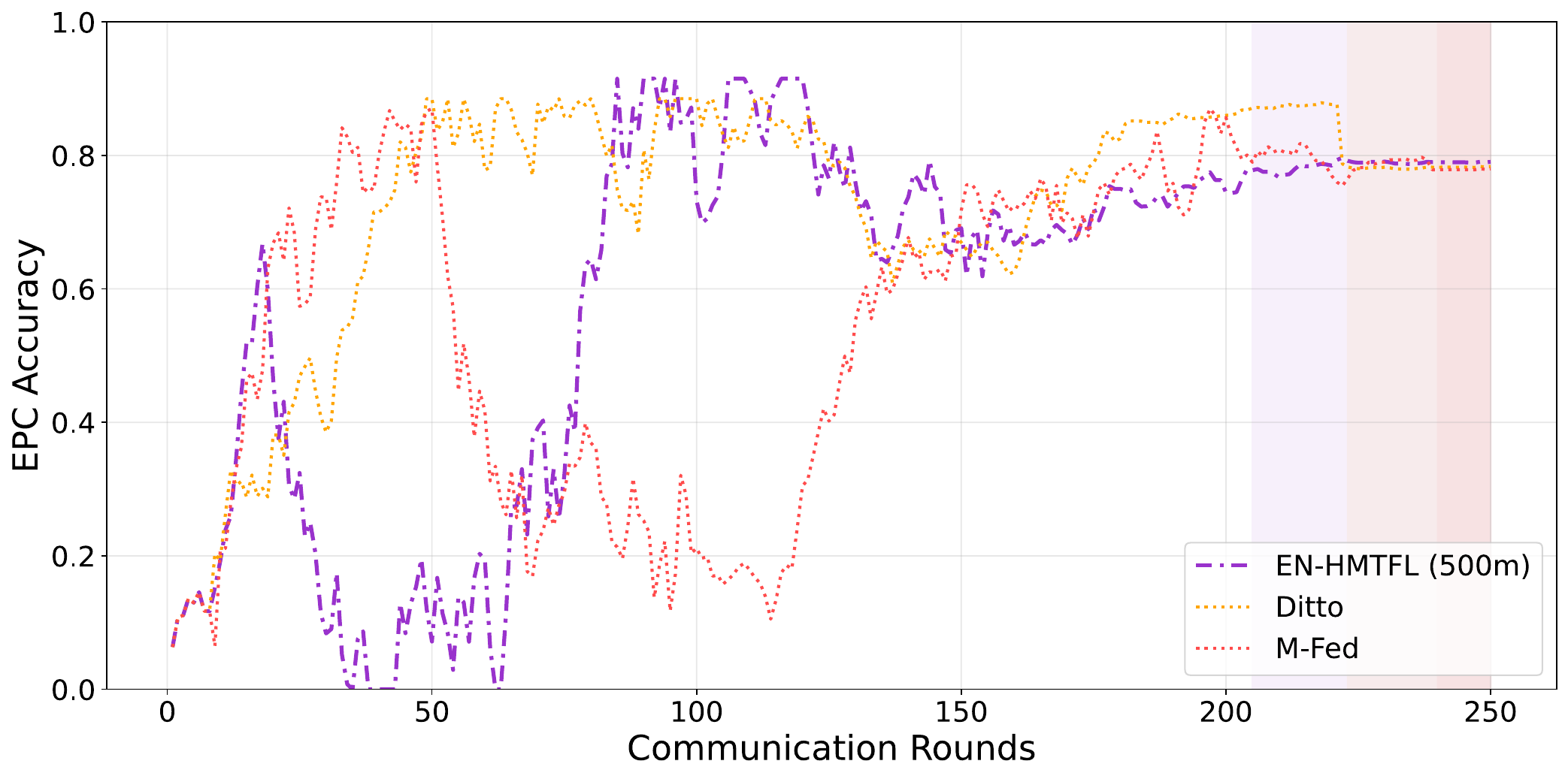}\label{fig:500_gtsrb}}
\caption{EPC accuracy versus communication round for 20 vehicles: (a)--(b) 100~m and (c)--(d) 500~m transmission ranges.}
\label{fig:range_all}
\end{figure*}

Fig.~\ref{fig:range_all} evaluates the effect of transmission range for 20 vehicles. For MNIST at 100~m, EN-HMTFL achieves a late-round mean EPC accuracy of $73.56\pm0.30\%$, compared with $72.07\pm0.37\%$ for Ditto and $63.71\pm0.38\%$ for M-Fed. This corresponds to improvements of 1.49 percentage points over Ditto and 9.85 percentage points over M-Fed, equivalent to relative gains of 2.1\% and 15.5\%, respectively. EN-HMTFL also converges at round 177, which is 51 rounds earlier than Ditto and 35 rounds earlier than M-Fed. This improvement results from combining cross-task encoder sharing with localized CH-level aggregation, which promotes transferable representation learning while limiting the propagation of heterogeneous updates under short-range connectivity.

At 500~m, EN-HMTFL remains the most accurate method for MNIST, reaching $72.82\pm0.33\%$. It exceeds Ditto by 0.75 percentage points and M-Fed by 9.10 percentage points, corresponding to relative gains of 1.0\% and 14.3\%, respectively. EN-HMTFL and Ditto satisfy the convergence criterion at the same round, whereas M-Fed converges 16 rounds earlier. However, M-Fed stabilizes at an accuracy that is 9.10 percentage points lower, showing that earlier convergence does not necessarily indicate a better learned model. Compared with the 100~m case, the wider transmission range reduces the EN-HMTFL accuracy by 0.75 percentage points and delays convergence by 51 rounds. This behavior is attributed to the denser neighborhood, which improves reachability but also introduces a broader and more heterogeneous set of encoder updates into uniform aggregation.

\begin{table*}[t]
\centering
\caption{EPC Accuracy and Convergence for the different Scenarios}
\label{tab:results}
\scriptsize
\setlength{\tabcolsep}{4.0pt}
\renewcommand{\arraystretch}{1.04}
\begin{tabular}{|c|c|c|c|c|c|}
\hline
\textbf{Scenario} & \textbf{Algorithm} & \textbf{MNIST Avg. Acc. (\%)} & \textbf{MNIST Conv. (round)} & \textbf{GTSRB Avg. Acc. (\%)} & \textbf{GTSRB Conv. (round)} \\ \hline
\multirow{3}{*}{\shortstack{50 vehicles\\100 m}}
& \textbf{EN-HMTFL} & \textbf{$69.29\pm0.22$} & 211 & \textbf{$76.39\pm7.79$} & 233 \\ \cline{2-6}
& Ditto & $61.89\pm0.67$ & \textbf{209} & $29.92\pm0.41$ & \textbf{211} \\ \cline{2-6}
& M-Fed & $49.81\pm1.06$ & 223 & $61.58\pm0.86$ & 238 \\ \hline
\multirow{3}{*}{\shortstack{20 vehicles\\100 m}}
& \textbf{EN-HMTFL} & \textbf{$73.56\pm0.30$} & \textbf{177} & \textbf{$81.81\pm0.07$} & \textbf{171} \\ \cline{2-6}
& Ditto & $72.07\pm0.37$ & 228 & $78.19\pm0.12$ & 223 \\ \cline{2-6}
& M-Fed & $63.71\pm0.38$ & 212 & $78.56\pm0.74$ & 240 \\ \hline
\multirow{3}{*}{\shortstack{20 vehicles\\500 m}}
& \textbf{EN-HMTFL} & \textbf{$72.82\pm0.33$} & 228 & \textbf{$78.93\pm0.10$} & \textbf{205} \\ \cline{2-6}
& Ditto & $72.07\pm0.37$ & 228 & $78.19\pm0.12$ & 223 \\ \cline{2-6}
& M-Fed & $63.71\pm0.38$ & \textbf{212} & $78.56\pm0.74$ & 240 \\ \hline
\end{tabular}
\end{table*}

A similar trend is observed for GTSRB. At 100~m, EN-HMTFL achieves $81.81\pm0.07\%$, outperforming Ditto and M-Fed by 3.62 and 3.25 percentage points, respectively. These correspond to relative improvements of 4.6\% over Ditto and 4.1\% over M-Fed. EN-HMTFL converges at round 171, which is 52 rounds earlier than Ditto and 69 rounds earlier than M-Fed. The very small late-round standard deviation further confirms that the proposed method reaches a stable high-accuracy regime. At 500~m, EN-HMTFL still achieves the highest mean accuracy, $78.93\pm0.10\%$, while converging 18 rounds earlier than Ditto and 35 rounds earlier than M-Fed. Although the accuracy margins narrow to 0.74 percentage points over Ditto and 0.38 percentage points over M-Fed, EN-HMTFL remains superior in both accuracy and convergence speed. The stronger gains at 100~m indicate that hierarchical encoder aggregation is most effective when localized clusters reduce update heterogeneity, whereas the wider range weakens this advantage by mixing updates from a more diverse set of vehicles.

Table~\ref{tab:results} summarizes these accuracy and convergence results. Overall, EN-HMTFL consistently achieves the highest late-round EPC accuracy across both tasks and both transmission ranges. Its convergence behavior is scenario-dependent: EN-HMTFL converges earlier than both benchmarks in the 20-vehicle GTSRB scenarios and in the 20-vehicle MNIST scenario at 100~m, matches Ditto at 500~m for MNIST, and converges slightly later than Ditto in the 50-vehicle scenario. These results show that the principal advantage of EN-HMTFL is its consistently higher accuracy, while convergence gains are obtained in several, but not all, evaluated scenarios.

\section{Conclusion}
This paper introduced EN-HMTFL, a hierarchical multi-task federated learning framework for dynamic VANETs in which vehicles exchange a shared encoder through a cluster-based CM--CH--EPC hierarchy while keeping task-specific decoders local. This design enables knowledge transfer across heterogeneous yet related tasks without aggregating incompatible task-specific model components. Across the evaluated vehicle-density and transmission-range settings, EN-HMTFL consistently achieved the highest late-round EPC accuracy, with relative accuracy gains of up to 24.0\% over M-Fed. Its convergence behavior varied across scenarios: EN-HMTFL converged earlier than the compared methods in several settings, achieving a maximum reduction of 69 communication rounds (28.8\%), whereas in other settings it converged at the same round as, or later than, one benchmark. These results demonstrate the accuracy benefit of hierarchical encoder sharing while showing that the convergence advantage depends on the vehicular scenario. Future work will extend the framework to more complex perception and multimodal tasks and investigate task-aware clustering and reliability-aware aggregation.

\bibliographystyle{ieeetr}
\bibliography{references}

\end{document}